\documentclass[conference]{IEEEtran}
\renewcommand\IEEEkeywordsname{Keywords}
\usepackage{graphicx} 
\usepackage{subfigure} %
\usepackage{amssymb} 
\usepackage{amsmath} 
\usepackage{multirow}

\usepackage{cite}
\usepackage{authblk}
\usepackage[colorlinks, citecolor=green]{hyperref}
\newcommand{\Mark}[1]{\textsuperscript{#1}} 
\newcommand*{\email}[1]{\texttt{#1}}

\begin{document}
%
\title{ Design of a Very Compact CNN Classifier for Online Handwritten Chinese Character Recognition Using DropWeight and Global Pooling}

\author{%
 Xuefeng Xiao, Yafeng Yang, Tasweer Ahmad, {Lianwen Jin}\Mark{*} and Tianhai Chang\\
 {\small{School of Electronic and Information Engineering, South China University of Technology, Guangzhou, China}}\\
 \email{{xiaoxuefengchina,yangyafeng17,tasveerahmad,lianwen.jin}@gmail.com}\\
}
\maketitle

\begin{abstract}
Currently, owing to the ubiquity of mobile devices, online handwritten Chinese character recognition (HCCR) has become one of the suitable choice for feeding input to cell phones and tablet devices. Over the past few years, larger and deeper convolutional neural networks (CNNs) have extensively been employed for improving character recognition performance. However, its substantial storage requirement is a significant obstacle in deploying such networks into portable electronic devices. To circumvent this problem, we propose a novel technique called DropWeight for pruning redundant connections in the CNN architecture. It is revealed that the proposed method not only treats streamlined architectures such as AlexNet and VGGNet well but also exhibits remarkable performance for deep residual network and inception network. We also demonstrate that global pooling is a better choice for building very compact online HCCR systems. Experiments were performed on the ICDAR-2013 online HCCR competition dataset using our proposed network, and it is found that the proposed approach requires only 0.57 MB for storage, whereas state-of-the-art CNN-based methods require up to 135 MB; meanwhile the performance is decreased only by 0.91\%.
\end{abstract}


\IEEEpeerreviewmaketitle

\section{Introduction}

\par Over the last five decades \cite{kimura1987modified,dai2007chinese}, handwritten Chinese character recognition (HCCR) has attracted considerable attention from researchers and has extensively been studied owing to the large number of character classes, similarity between characters, and variation in writing style. Handwriting recognition can broadly be categorized into online and offline handwriting recognition. The main motivation of this paper is to deal with the problem of storage capacity for online HCCR by using some novel techniques such as GoogleNet. In contrast to offline HCCR, in which gray-scale images are analyzed and classified into different groups, for online HCCR, pen trajectories are the main source of information to recognize different characters \cite{TpamiLiuJN04}. Moreover, online HCCR finds numerous applications in pen input devices, personal digital assistants, smart phones, touch-screen devices, etc.

\par Academic and commercial research in HCCR has greatly progressed owing to handwriting recognition competitions held over the past few years \cite{liu2010chinese,liu2011icdar,yin2013icdar}. In these competitions, many participants began to use methods based on convolutional neural networks (CNNs) for HCCR, instead of conventional machine learning tools such as the MQDF-classifier \cite{kimura1987modified}. It was also demonstrated that methods based on CNNs can learn more discriminative representations from raw data and can lead to end-to-end solutions for HCCR problems. For the ICDAR 2013 online HCCR competition dataset \cite{yin2013icdar}, the novel DropSample training method was proposed in \cite{yang2016dropsample}, which achieved an accuracy of 97.23\% and subsequently of 97.51\% when ensembling nine model. Zhang et al. \cite{zhang2017online} combined conventional normalization-cooperated direction-decomposed feature maps and CNNs to achieve an accuracy of 97.55\% and of 97.64\% by voting on three models. Thus far, the state-of-the-art CNN-based architecture has produced an accuracy of 97.79\% \cite{LaiJY17} by using the DropDistortion training strategy.

\par Currently, CNN-based methods are quite popular to deal with the problems of character recognition, and it seems intuitive that the deployment of CNN in portable devices would improve the performance of online handwriting recognition. However, they demand a considerable amount of storage and memory bandwidth. For online HCCR, the aforementioned state-of-the-art methods \cite{yang2016dropsample,zhang2017online,LaiJY17} require storage spaces of 135.0 MB, 70.50 MB, and 19.03 MB, respectively. This requirement of large storage space is the main hindrance in deploying such deep networks into portable devices such as mobile phones. The large and deep networks are not a pragmatic choice for on-chip storage as they demand additional memory resources. This problem has led to the proposal of some compact designs that would be viable to deploy in portable electronic gadgets.

\begin{figure*}[htbp]
\centering

\subfigure[]{
\label{distortionImage}
\includegraphics[width=0.95\textwidth]{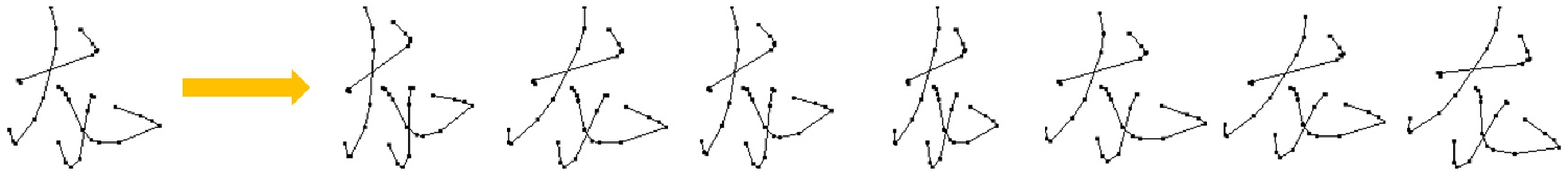}}

\subfigure[]{
\label{pathSignature}
\includegraphics[width=0.95\textwidth]{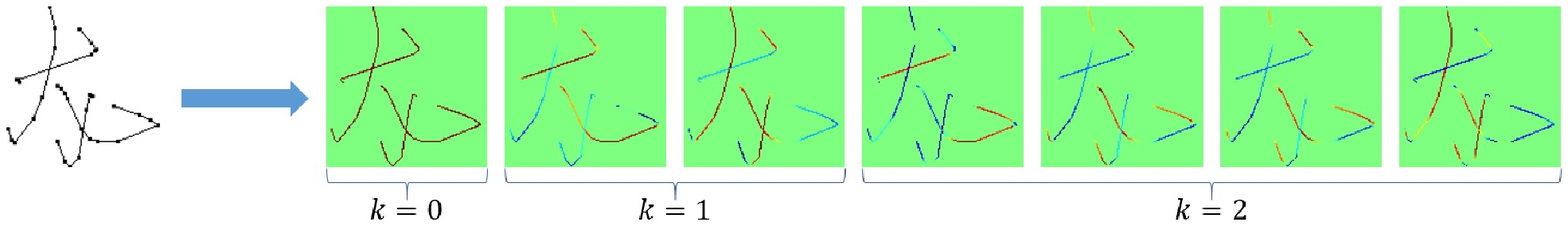}}

\caption{(a) Characters distortion. (b) Path signature feature map of input pen-tip trajectories.}
\label{fig:2}
\end{figure*}

\par Recently, many researchers have attempted to build compact networks. Prominently, network pruning \cite{han2015learning,guo2016dynamic} is the most effective method to compress CNNs by pruning the redundant connections in each layer. However, to our knowledge, no study has investigated whether these methods are feasible for large-scale online HCCR involving more than 3,700 classes of characters. In previous \cite{XiaoJYYSC17}, they adopted the network pruning technique to compress a model built for offline HCCR to 2.3 MB. In the present paper, we propose the use of the DropWeight technique for online HCCR. We also reveal that the DropWeight technique is immune to network architectures and that it can be applied to compress various types of deep network structures, such as VGGNet \cite{Simonyan2014VeryDC}, GoogLeNet \cite{szegedy2015going}, and ResNet \cite{He_2016_CVPR}. In the present work, we also demonstrate that the use of global pooling is a good choice for building compact online HCCR systems. We conducted an experiment with the ICDAR 2013 online HCCR competition dataset, based on which we carefully designed deeper and thinner networks that use global pooling, which only costs 9.9 MB storage before connections pruning. After integrating our DropWeight technique, the model costs only 0.57 MB storage. The accuracy of this model is slightly lower than that of the state-of-the-art CNN-based network by 0.91\%, but it costs only 1/33 of the storage required for the best CNN model so far reported in the literature.

\par The remainder of this paper is organized as follows. Section II gives a brief description of the DropWeight technique. Section III highlights various contemporary network architectures. Experiments and results are presented in section IV, and section V concludes the results.

\section{DropWeight}

\par As CNN architectures are very large in size, it is quite desirable to compress the networks by pruning redundant connections. Therefore, the network connections of a network trained in advance are pruned with the idea that weights lower than a threshold should be removed, thereby converting a dense, fully-connected network to a sparse network \cite{han2015learning}. The network is pruned and retrained iteratively so that its performance does not degrade significantly.

\par As proposed by \cite{han2015learning,guo2016dynamic}, the fixed pruning threshold is computed as follows:
\begin{equation}
\begin{aligned}
{{P_{th}} = \frac{\eta }{N}\sum\limits_{i = 1}^N {\left| {{w_i}} \right|}  + \beta \sqrt {\frac{1}{N}\sum\limits_{i = 1}^N {(w{}_i - } \frac{1}{N}\sum\limits_{i = 1}^N {w{}_i} {)^2}}  + \lambda }.
\end{aligned}
\end{equation}

\par For the layer containing $N$ weights, the pruning threshold ${P_{th}}$ is mainly dependent on the average absolute value and variance of weights of layer ${w_i}$, but the value of this threshold can also be empirically determined by varying the parameters $\eta,\beta$ and $\lambda$. However, the application of a fixed threshold is complicated, as an excessively high threshold would remove many significant connections at the start, and it will be very difficult for the network to recover the original performance; conversely, if the threshold is too low, the desired compression ratio may not be achieved.

\par To address this problem, we adopt the DropWeight technique, in which the threshold is gradually increased. In experiments, we prune the connections after every $I$ iterations (in experiments, we set $I = 10$). If we wish to prune a certain percentage of connections for a layer, the pruning number must be increased for each pruning iteration. Therefore, the threshold is determined by the pruning number. The absolute values of weights below this threshold are set to zero. By dynamically increasing the pruning number, the threshold would also be gradually increased. During iterations without the pruning process, the weights are updated with a gradient, and the pruned weights cannot be retrieved. Once the desired pruning ratio is reached, the increasing threshold is fixed and noted for further pruning of the layer until pruning ends. Finally, the weight quantization topology, as proposed by \cite{Han2015DeepCC}, is incorporated, following which this quantized-pruned network is fine-tuned for improving performance.

\section{Network Architectures for online HCCR}

\subsection{Characters Distortion}

\par A potential problem in online character recognition is the variation in handwriting style. To address this problem, the concept of character distortion is introduced to generate a large number of training samples artificially. Character distortion is produced by introducing an affine transformation and its variants to the training samples \cite{yang2015improved}. In character distortion, a nonlinear normalization, as proposed by \cite{leung2009recognition}, is also entertained for character shearing and stroke distortion.

\par Let $\alpha$ be the total degree of character distortion, $\theta$ a number ranging from ($-\alpha$, $\alpha$), $[x, y]$ the pixel coordinates before transformation, and $[x¡ä, y¡ä]$ the pixel coordinates after transformation. Then, affine transformations are formulated as follows:

\begin{equation}
\label{distort1}
\begin{aligned}
{[x',y'] \Leftarrow [x,y] \cdot \left[ {\begin{array}{*{20}{c}}{1 + {\alpha _x}}&0\\0&{1 + {\alpha _y}}\end{array}} \right],}
\end{aligned}
\end{equation}

\begin{equation}
\label{distort2}
\begin{aligned}
{[x',y'] \Leftarrow [x,y] \cdot \left[ {\begin{array}{*{20}{c}}1&\alpha \\0&1\end{array}} \right],}
\end{aligned}
\end{equation}

\begin{equation}
\label{distort3}
\begin{aligned}
{[x',y'] \Leftarrow [x,y] \cdot \left[ {\begin{array}{*{20}{c}}1&0\\\alpha &1\end{array}} \right],}
\end{aligned}
\end{equation}

\begin{equation}
\label{distort4}
\begin{aligned}
{[x',y'] \Leftarrow [x,y] \cdot \left[ {\begin{array}{*{20}{c}}{\cos \left( \alpha  \right)}&{ - \sin \left( \alpha  \right)}\\
{\sin \left( \alpha  \right)}&{\cos \left( \alpha  \right)}
\end{array}} \right],}
\end{aligned}
\end{equation}
Eq.\ref{distort1} and Eq.\ref{distort2} tilts the strokes; Eq.\ref{distort3} stretches or shrinks strokes, and Eq.\ref{distort4} generates rotational distortion.

\par As shown in Fig.\ref{distortionImage}, global character stretching, scaling, rotation, and translation are performed using affine transformations, whereas local distortion is performed using one-dimensional deformation and non-linear normalization, as proposed by \cite{leung2009recognition}. Non-linear normalization produces character shearing and stroke distortion.

\subsection{Path Signatures}

\par The idea of path signatures was originally proposed by Chen et al. \cite{chen1958integration} as an iterated integral for solving differential equations. This concept of path signatures was implemented as a set of features by \cite{Graham13} to improve the performance of CNNs for online handwritten character recognition. Empirically, it is revealed that the first and second iterated integrals entail significant information for CNNs. Mathematically, for positive integers k and intervals $[s, t]$, the k-th iterated integral of X is the ${d^k}$-dimensional vector defined by

\begin{equation}
\begin{aligned}
{I_{s,t}^k = \int_{s < {u_1} < ... < {u_k} < t} {1d{X_{{u_1}}} \otimes } ... \otimes d{X_{{u_k}}},}
\end{aligned}
\end{equation}

where $\otimes$ denotes the tensor product. For k=0, the iterated integral is 1 and corresponds to its offline image; for k=1, the iterated integral corresponds to path displacement; and for k=2, the iterated integral corresponds to path curvature, as shown in Fig.\ref{pathSignature}.

\begin{figure} \centering
\subfigure[] {
\label{streamline}
\includegraphics[height=.7\textwidth]{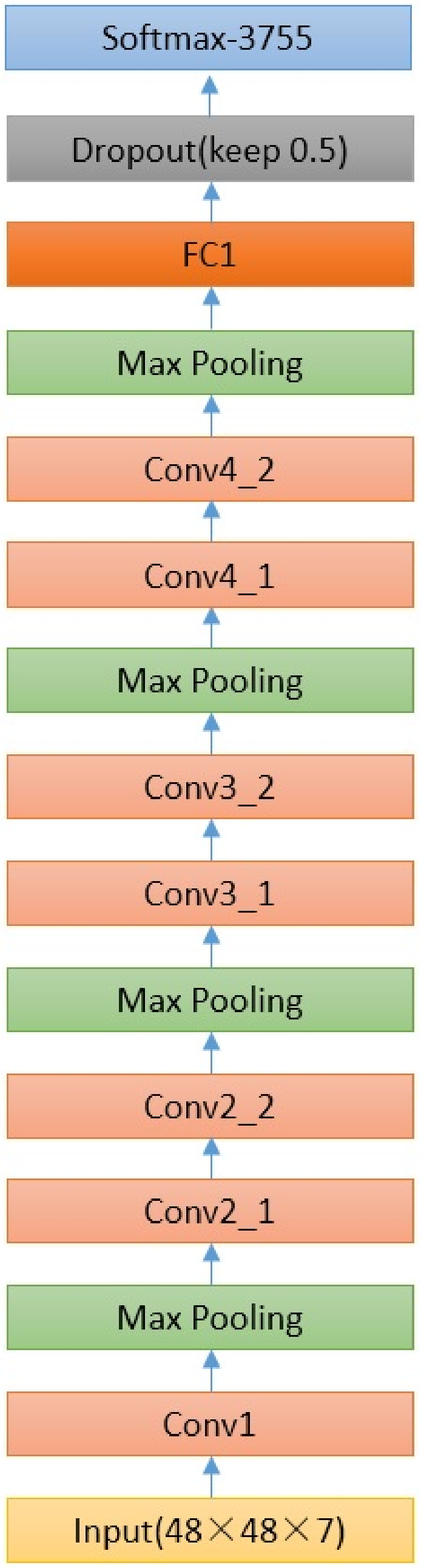}
}
\subfigure[] {
\label{inception}
\includegraphics[height=.7\textwidth]{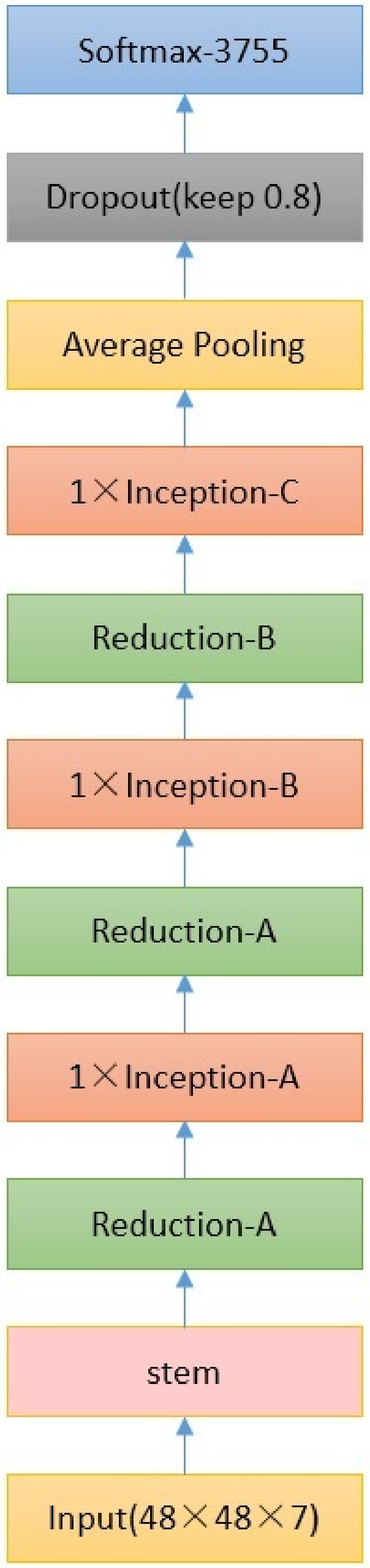}
}
\caption{Network structure of (a) HCCR-Str-FC and (b) HCCR-Inc-GAP. }
\label{fig}
\end{figure}

\subsection{Network Structure}

\par For online HCCR, we designed three different network structures. The first is a conventional streamlined CNN, as shown in Fig.\ref{streamline}. In this network, all convolutional filters were $3\times3$ size with one padding pixel for retaining the original size. A max-pooling operation was performed over a $3\times3$ window with a stride of 2. All convolutional layers and the first fully connected layer were equipped with a batch normalization (BN) \cite{ioffe2015batch} layer, and PReLU \cite{He2015DelvingDI} was added to each BN layer. The overall architecture can be represented as Input-128C3-MP3-160C3-160C3-MP3-256C3-256C3-MP3-384C3-384C3-MP3-1024FC-Output. Then, to reduce the original network's storage size, we use a global average pooling (GAP) layer to replace the last pooling layer and the first fully connected layer. We refer to the two networks as HCCR-Str-FC and HCCR-Str-GAP, respectively.

\par The second network structure we used is residual network \cite{He_2016_CVPR}, which introduces the short-cut connections to smoothly pass the gradients into shallow layers for solving the problem of vanishing gradients. This architecture won the first place in the ILSVRC 2015 classification challenge. We used an 18-layer architecture in which the output channels of all convolution layers were decreased by 50\%. For comparison, we used a fully connected layer that contains 1024 neurons to replace the original global pooling layer. We refer to the two networks as HCCR-Res-FC and HCCR-Res-GAP, respectively.

\par The last network structure we used is the Inception-v4 \cite{SzegedyIVA17} network. It is mainly based on GoogLeNet or Inception-v1 \cite{szegedy2015going}, which was introduced to address the challenges of memory utilization and computational cost. In order to make it suitable for online HCCR, we removed the first two convolutional layers in the stem module and added a Reduction-A block between the stem block and Inception-A block. As shown in Fig.\ref{inception}, we only used three inception modules, and the output channels of all convolution layers were decreased in size by 75\%. For fair comparison, we used a fully connected layer that contains 1024 neurons to replace the original global pooling layer. We refer to the two networks as HCCR-Inc-FC and HCCR-Inc- GAP, respectively.

\section{Experiment}

\subsection{Network Training}

\par Our proposed framework was evaluated using online HCCR dataset. The network was trained using the OLHWDB1.0 and OLHWDB1.1 datasets \cite{liu2011casia}, and the performance of the proposed network was tested using the On-ICDAR2013 dataset \cite{yin2013icdar}, which contained 3,755 classes. The network was trained using a set of 2,693,183 samples from 720 different subjects, whereas it was evaluated using 224,590 test images from 60 different writers.

\par For training the online HCCR network, the distortion technique was used for each sample at each training epoch. Path signature feature maps were extracted from online handwritten Chinese characters, and these feature maps were fed as the input for training the network. The baseline model was trained on the Caffe \cite{jia2014caffe} deep learning platform with a mini-batch size of 128 and momentum of 0.9. The learning rate was initialized with 0.1, and it was reduced by 0.1 after every 70,000 iterations. The training process concluded after 300,000 iterations.

\begin{table}[htbp]
\centering
\caption{Compression results for different network structures}
\begin{tabular}{l|c|c|c|c}
\hline
\multirow{2}{*}{Model} &  \multicolumn{2}{|c|}{Before Compression}  &  \multicolumn{2}{|c}{After Compression}  \\
\cline{2-5}
&Stor.(MB)&Accu.(\%)&Stor.(MB)&Accu.(\%)\\
\hline
HCCR-Str-GAP&19.24 & 97.51&  0.84 & 96.62\\
HCCR-Str-FC&41.93 & 97.77&  1.18 & 96.49\\
\hline
HCCR-Res-GAP&14.40 & 96.89&  0.70 & 96.05\\
HCCR-Res-FC&29.42 & 97.02&  1.07 & 96.03\\
\hline
HCCR-Inc-GAP&9.36 & 97.45&  0.57 & 96.88\\
HCCR-Inc-FC&56.05 & 97.65&  0.76 & 96.83\\
\hline
\end{tabular}
\label{compressSixModel}
\end{table}

\subsection{Accuracy and Storage}

\par Tab.\ref{compressSixModel} presents a comparative analysis of storage and accuracy for the ICDAR-2013 online competition database for our six proposed networks. For the same structure, we can find that the use of global pooling to replace the fully connected layer slightly decreases the performance but significantly decreases the storage space required. Therefore, by using the DropWeight technique for the six networks, the storage capacity is drastically decreased, whereas the accuracy is only slightly decreased. For the streamlined, residual, and inception-based network, it is initially observed that the global pooling layer achieves a slightly lower accuracy compared to that of the fully connected layer. However, after compression, the performance of global pooling networks is better than those of fully connected layer networks; in addition, the storage required for the former is lower than that required for the latter. Thus, it is clearly demonstrated that global pooling is a good choice to build a compact system for online HCCR.

\par Tab.\ref{resultCompare} illustrates the results of three previous CNN-based methods \cite{liu2010chinese,liu2011icdar,yin2013icdar} that have achieved the highest performance thus far on the ICDAR-2013 online database. It is clear that our HCCR-Inc-GP can achieve a very compact design as compared with the three previous architectures, costing only 9.9 MB of memory. Moreover, after further compression, it costs merely 0.57 MB of memory and can still reach an accuracy of 96.88\%, which is certainly higher but requires 210 times smaller storage compared to the conventional method (DFE + DLQDF) \cite{LiuYWW13}. Even compared with the state-of-the-art CNN models for online HCCR, our model is 1/33 times more cost efficient while the performance is decreased only by 0.91\%.

\begin{table}[htbp]
\centering
\caption{Result for ICDAR-2013 online HCCR competition dataset}
\begin{tabular}{l|c|c|c}
\hline
Method & Ref. &Storage(MB)  & Accuracy(\%)\\
\hline
Traditional Method: DFE+DLQDF& \cite{LiuYWW13} & 120.0 & 95.31 \\
\hline
DropSample& \cite{yang2016dropsample} & 135.0 & 97.51 \\
\hline
DirectMap+ConvNet& \cite{zhang2017online} & 70.50 & 97.64\\
\hline
DropDistortion& \cite{LaiJY17} &  \textbf{19.03} & \textbf{97.79}\\
\hline
HCCR-Inc-GAP& ours&  9.90 & 97.45\\
\hline
HCCR-Inc-GAP-Pruned& ours&  \textbf{0.57} & \textbf{96.88}\\
\hline
\end{tabular}
\label{resultCompare}
\end{table}

\section{Conclusion}

\par In this paper, we proposed the DropWeight technique to compress popular CNN architectures for online HCCR, which includes a streamlined, residual, and inception-based network. We also demonstrated that global pooling is a good choice to build a compact network for online HCCR. Finally, we built a network that costs only 0.57 MB of storage but can still achieve an accuracy comparable to those of state-of-the-art CNN models. In the future, we will extend the method to other deep learning model such as long short-term memory (LSTM) network to address the problem of online handwritten text recognition and natural language processing.


\end{document}